\documentclass[sigconf]{acmart}

\AtBeginDocument{%
  \providecommand\BibTeX{{%
    \normalfont B\kern-0.5em{\scshape i\kern-0.25em b}\kern-0.8em\TeX}}}

\copyrightyear{2022}
\acmYear{2022}
\setcopyright{rightsretained}
\acmConference[SIGIR '22]{Proceedings of the 45th International ACM SIGIR Conference on Research and Development in Information Retrieval}{July 11--15, 2022}{Madrid, Spain}
\acmBooktitle{Proceedings of the 45th International ACM SIGIR Conference on Research and Development in Information Retrieval (SIGIR '22), July 11--15, 2022, Madrid, Spain}
\acmDOI{10.1145/3477495.3531954}
\acmISBN{978-1-4503-8732-3/22/07}

\usepackage{graphicx}
\usepackage{caption}
\usepackage{times}  
\usepackage{helvet}  
\usepackage{courier}  
\usepackage[noend]{algpseudocode}
\usepackage{algorithm}
\usepackage{algorithmicx}
\usepackage{threeparttable}
\usepackage{multirow}
\usepackage{subfigure}
\usepackage{dsfont}
\usepackage{booktabs}
\usepackage{microtype}
\usepackage{hyperref}
\usepackage{stfloats}

\usepackage{etoolbox}

\makeatletter
\patchcmd{\maketitle}{\@copyrightpermission}{
   \begin{minipage}{0.3\columnwidth}
     \href{http://creativecommons.org/licenses/by/4.0/}{\includegraphics[width=0.70\textwidth]{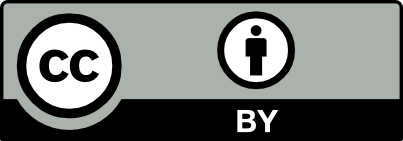}}
   \end{minipage}\hfill
   \begin{minipage}{0.7\columnwidth}
     \href{http://creativecommons.org/licenses/by/4.0/}{This work is licensed under a Creative Commons Attribution International 4.0 License.}
   \end{minipage}
  
}{}{}

\makeatother

\newcommand{\modelName}{ProbExpan}

\begin{document}

\fancyhead{}
\title{Contrastive Learning with Hard Negative Entities for \\Entity Set Expansion}

\author{Yinghui Li}
\authornote{These authors contributed equally to this research}
\email{liyinghu20@mails.tsinghua.edu.cn}
\orcid{0000-0001-7571-6722}

\affiliation{%
  \institution{Tsinghua Shenzhen International Graduate School, Tsinghua University}
  \city{Shenzhen}
  \state{Guangdong}
  \country{China}
}

\author{Yangning Li}
\authornotemark[1]
\email{liyn20@mails.tsinghua.edu.cn}
\orcid{0000-0002-1991-6698}
\affiliation{%
  \institution{Tsinghua Shenzhen International Graduate School, Tsinghua University}
  \city{Shenzhen}
  \state{Guangdong}
  \country{China}
}

\author{Yuxin He}
\authornotemark[1]
\email{21S051047@stu.hit.edu.cn}
\orcid{0000-0003-1105-0014}
\affiliation{%
  \institution{Harbin Institute of Technology}
  \city{Shenzhen}
  \state{Guangdong}
  \country{China}
}

\author{Tianyu Yu}
\orcid{0000-0001-9752-6655}
\affiliation{%
  \institution{Tsinghua Shenzhen International Graduate School, Tsinghua University}
  \city{Shenzhen}
  \state{Guangdong}
  \country{China}
}

\author{Ying Shen}
\orcid{0000-0002-3220-904X}
\authornote{Corresponding author: sheny76@mail.sysu.edu.cn, zheng.haitao@sz.tsinghua.edu.cn.}
\affiliation{%
  \institution{School of Intelligent Systems Engineering, Sun-Yat Sen University}
  \city{Shenzhen}
  \state{Guangdong}
  \country{China}
}

\author{Hai-Tao Zheng}
\orcid{0000-0001-5128-5649}
\authornotemark[2]
\affiliation{%
  \institution{Tsinghua Shenzhen International Graduate School, Tsinghua University}
  \city{Shenzhen}
  \state{Guangdong}
  \country{China}
}


\begin{abstract}
Entity Set Expansion (ESE) is a promising task which aims to expand entities of the target semantic class described by a small seed entity set.
Various NLP and IR applications will benefit from ESE due to its ability to discover knowledge. 
Although previous ESE methods have achieved great progress, most of them still lack the ability to handle \emph{hard negative entities} (i.e., entities that are difficult to distinguish from the target entities), since two entities may or may not belong to the same semantic class based on different granularity levels we analyze on. 
To address this challenge, we devise an entity-level masked language model with contrastive learning to refine the representation of entities. 
In addition, we propose the \modelName{}, a novel probabilistic ESE framework utilizing the entity representation obtained by the aforementioned language model to expand entities. 
Extensive experiments\footnote{The source codes are available at \url{https://github.com/geekjuruo/ProbExpan}.} and detailed analyses on three datasets show that our method outperforms previous state-of-the-art methods. 
\end{abstract}



\maketitle

\section{Introduction}
\label{sec:intro}

Entity Set Expansion (ESE) aims to expand from a set of seed entities (e.g., {“\emph{China}”, “\emph{America}”, “\emph{Japan}”}) to more new target entities (e.g., {“\emph{Russia}”, “\emph{Germany}”, ...}) that belong to the same semantic class (i.e., \texttt{Country}) as the seed entities. 
The ESE task can benefit kinds of NLP or IR downstream applications, such as knowledge graph construction~\citep{shi2021entity}, Web search~\citep{chen2016long}, and question answering~\citep{wang2008automatic}.

\begin{figure}[h]
\centering
\includegraphics[width=0.48\textwidth]{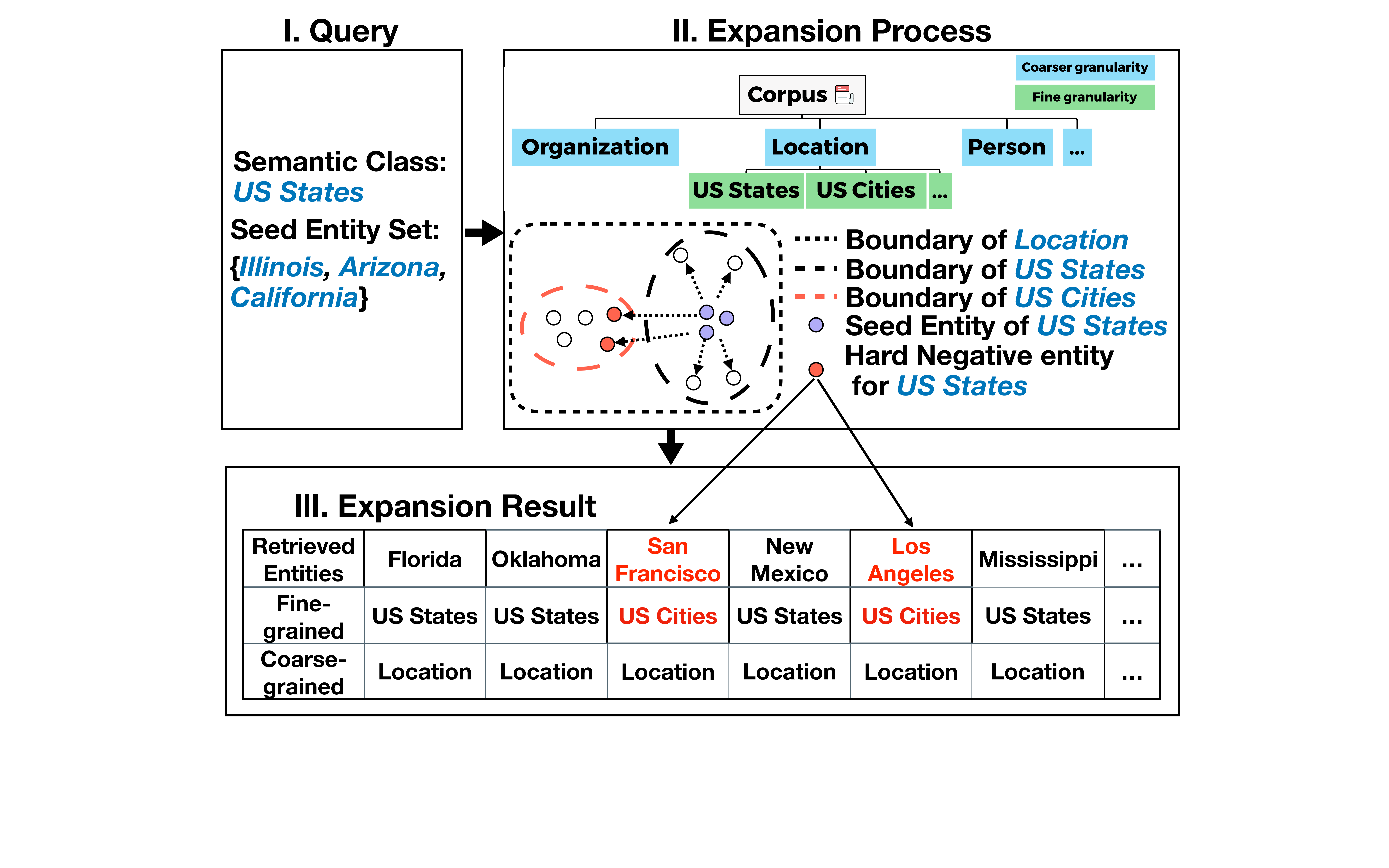}
\caption{Examples of hard negative entities in ESE.}
\label{Intro_Figure}
\end{figure}

In recent years, kinds of iterative bootstrapping methods have gradually become the mainstream of ESE research. 
These methods~\citep{shen2017setexpan, yan2019learning, CaSE} mainly select the most confident candidate entities of the model to the expanded set iteratively. 
A core challenge for these methods is to avoid selecting \emph{hard negative entities} that are semantically ambiguous with the target entities~\citep{jindal2011learning, gupta-manning-2014-improved}. 
As shown in Figure~\ref{Intro_Figure}, when we want to expand target entities belong to class \texttt{US States}, a competitive model is likely to wrongly expand hard negative entities, such as “\emph{San Francisco}” and “\emph{Los Angeles}”. If judged according to \texttt{US States}, it is clear that these hard negative entities do not belong to the target semantic class. But if we follow a coarser granularity (i.e., \texttt{Location}), hard negative entities can be regarded as the same class as target entities. 
Furthermore, as the characteristic of the iterative expansion process, a small number of negative entities that are incorrectly expanded in the early iterations will cause errors to accumulate in later iterations, resulting in a gradual decrease in expansion performance. This is the long-term “semantic drift” problem faced by ESE methods~\citep{curran2007minimising, mcintosh-2010-unsupervised, shi-etal-2014-probabilistic}.

To address the above challenge and problem, we propose to use contrastive learning~\citep{chen2020simple, robinson2020hard, gao2021simcse} to empower the ESE model to better deal with hard negative entities. 
Contrastive learning is a hot topic in self-supervised learning field, which is originally applied to learn visual representation and attracts more and more attention from NLP researchers. 
In general, contrastive learning aims to learn more effective representation by pulling samples belonging to similar classes and further pushing samples from different classes~\citep{HadsellCon}.
Intuitively, the idea that we want the ESE model to better distinguish hard negative entities coincides with the motivation of contrastive learning naturally.
In our study, contrastive learning provides the ESE model with clearer semantic boundary and make entity representation closer to semantics.  

Motivated by the above intuition, we propose a novel ESE method that consists of three parts: 
(1) \emph{Entity representation model}, an entity-level masked language model pre-trained under the entity prediction task we specially design for ESE. Then we apply contrastive learning to refine the semantic representation learned by our model, which can be utilized in later expansion process.
(2) \emph{Model selection and ensemble}. Due to the randomness of training samples in the pre-training process, the model will be sensitive to the quality of the training context features.
To alleviate this issue, we will pre-train multiple models mentioned in (1), then select and ensemble top models to avoid the randomness of single model. 
(3) \emph{Probabilistic expansion framework}, a novel framework that can utilize the ensemble model obtained in (2) through a window search algorithm and an entity re-ranking algorithm, both based on probabilistic representation similarity of the candidate entity and entity set. Through this framework, we can finally get the ideal target entities that we want to expand.

In summary, our contributions are in three folds:
\begin{itemize}
    \item We firstly apply contrastive learning in ESE to better handle hard negative entities and derive more effective entity representation in semantic space. 
    \item We propose a novel ESE framework, \modelName{}, which can uniformly represent entity/entity set in the probability space and utilize the ESE model to expand target entities. 
    \item We conduct extensive experiments and detailed analysis on three public datasets and get state-of-the-art performance. Solid results demonstrate the substantial improvement of our method over previous baseline methods.
\end{itemize}

\section{Related Work}
\label{sec:rw}
\noindent\textbf{Entity Set Expansion.} 
Recently, many corpus-based ESE methods have gradually become the mainstream paradigm. These corpus-based ESE methods can be divided into two main categories:(1) one-time ranking methods~\citep{mamou-etal-2018-term,CaSE,kushilevitz-etal-2020-two} which introduce pairwise semantic similarity into set expansion tasks and suffer a lot from the \emph{Entity Intrusion Error} problem, that is, they cannot clearly convey the semantic meaning of entities. (2) iterative pattern-based bootstrapping methods~\citep{Egoset,shen2017setexpan, AuxiliaryExpan} which aim to bootstrap the seed entities set by iteratively selecting context pattern and ranking expanded entities. But these methods usually are troubled by the \emph{Semantic Drift} problem, that is, the target semantic class will change gradually when noise arises during the iterations. 

\noindent\textbf{Language Representation.} 
Early representation methods focus on word-level embeddings, such as Word2Vec~\citep{goldberg2014word2vec} and Glove~\citep{pennington2014glove} which output a single embedding for each word in vocabulary. After that, researchers design many context-aware representation methods to utilize context information better. The outstanding representative of context-aware representation methods is the masked language models represented by BERT~\citep{devlin-etal-2019-bert}. It is noted that CGExpan~\citep{zhang-etal-2020-empower} has utilized BERT's representation to enhance ESE. But BERT can still only perform word-level representation and CGExpan only use the pre-trained BERT embeddings without any task-specific training. 
To obtain better representation in more complex tasks, ERNIE~\citep{zhang-etal-2019-ernie} is designed to learn entity/phrase-level representation. To the best of our knowledge, entity-level representation methods have not yet been applied to ESE. 

\noindent\textbf{Contrastive Learning.}
Contrastive learning has been widely applied in self-supervised field~\citep{kim-etal-2021-self, qin-etal-2021-erica, wang-etal-2021-cline,li2022past}.  The main motivation of contrastive learning is to attract the positive samples and repulse the negative samples~\citep{HadsellCon,chen2020simple,khosla2020supervised,gao2021simcse}. 
Recent work~\citep{robinson2020hard} shows that contrastive representation learning benefits from hard negative samples(those samples which are difficult to distinguish from positive samples). This idea coincides with the challenge we have observed in the existing ESE methods, that is, most expansion models cannot handle hard negative entities well. 
SynSetExpan~\citep{shen-etal-2020-synsetexpan} enhances the ESE task via another related task, Synonym Discovery. It is different from the idea of contrastive learning, hoping to suppress the effect of negative entities by obtaining more positive entities.
NEG-FINDER~\citep{mcintosh-2010-unsupervised} is concerned with semantic negative classes same as our work, but it proposes to perform offline negative discovery and then utilize the pre-selected negative categories to alleviate the semantic drift of the bootstrapping algorithms. 
Unlike NEG-FINDER that is just a heuristic and untrainable algorithm, our study aims to pre-train a task-specific model which has better entity representation and clearer semantic boundaries for ESE by contrastive learning.

\section{Methodology}
\label{sec:method}

\begin{figure*}[]
\centering
\includegraphics[width=1.00\textwidth]{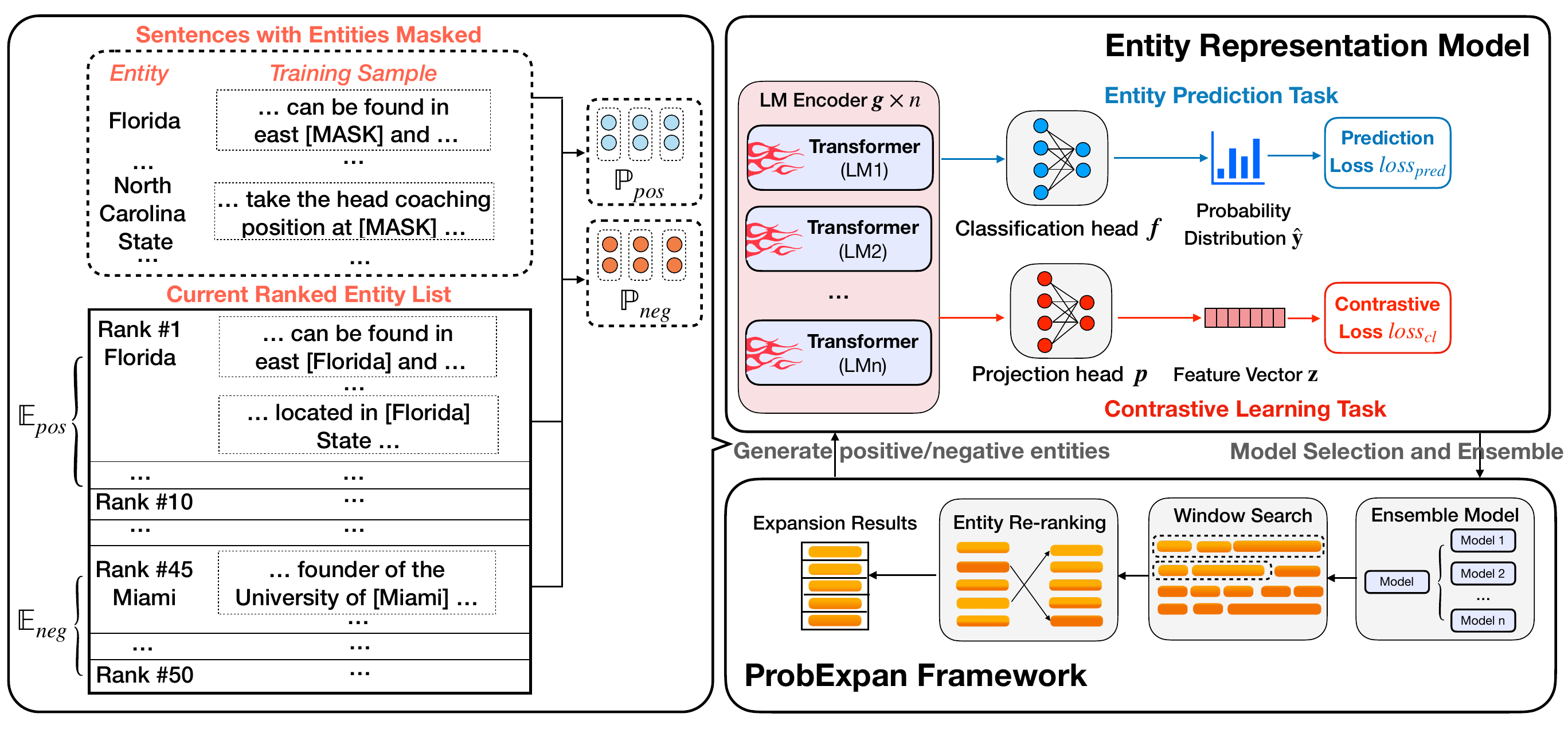}
\caption{Overview of our proposed method. We jointly train the entity representation model to obtain clearer semantic boundaries through our designed entity prediction task and contrastive learning task. Based on multiple pre-trained entity representation models, we utilize the model selection and ensemble mechanism to avoid the randomness brought by a single model. Two simple yet effective algorithms, namely window-search and entity re-ranking algorithms, are used to search and sort entities to obtain ideal target entities, according to the similarity of probabilistic representations derived from the ensemble model.}
\label{Method_Figure}
\end{figure*}

In this section, we firstly introduce the entity representation model and the entity prediction task we design for ESE. Specially, we will discuss how we apply contrastive learning to refine entity representation. Then we will illustrate the mechanism of model selection and ensemble. Finally, we will describe the expansion framework and algorithm to expand target entities. The overview of our proposed method is shown in Figure~\ref{Method_Figure}.

\subsection{Entity Representation Model}
\label{sec:methodmlm}
The entity representation model mainly contains an entity-level masked language model, which takes a tokenized sentence with entity masked as input and outputs a probability distribution describing which entity the masked token can be. 
The entity representation is defined as the average of the predicted entity distributions of all its sentences. 
And the representation of an entity set is defined as the average of the representation of all its entities.
Our entity-level masked language model contains an encoder $\boldsymbol{g}$ and a classification head $\boldsymbol{f}$. 
To be specific, we initialize the encoder $\boldsymbol{g}$ with pre-trained parameters of $\text{BERT}_\text{BASE}$, so that grammatical and semantic knowledge learned from large-scale corpus by BERT can be utilized. 
The classification head $\boldsymbol{f}$ consists of two linear layers with GeLU activation and a softmax layer.
We set biases of the classification head $\boldsymbol{f}$ to be 0 and initialize weights from Kaiming uniform distribution~\citep{He_2015_ICCV}.

Concerning the masked entity prediction pre-training task, for every entity in the vocabulary, we replace its span with [MASK] to get a training sample for all the sentences that it appears. 
During each training epoch, we restrict the number of samples from every entity to be the average number of samples of all entities, for concern of sample imbalance and the facility of following ensemble learning. We choose the Label Smoothing loss function~\citep{szegedy2016rethinking} rather than traditional cross-entropy loss function, so that entities sharing similar semantic meaning with target entity will not be overly suppressed. The prediction loss is defined as:
\begin{equation}
\begin{aligned}
    loss_{pred} = - \frac{1}{N} \sum_i^N \sum_j^{V_e} (&\mathds{1}_{j=y_i}(1-\eta)\cdot \log \hat{\mathbf{y}}_i[j] \\&+ \mathds{1}_{j\neq y_i}\eta \cdot \log \hat{\mathbf{y}}_i[j]),
\end{aligned}
\end{equation}
where $N$ is the size of mini-batch, $V_e$ is the size of entity vocabulary, $\eta$ is the smoothing factor and the larger the $\eta$ is, the higher the smoothness of the label is, $y_i$ is the index of the entity corresponding to the training sample $i$, $\hat{\mathbf{y}}_i$ is the output of $\boldsymbol{f}$.

\subsection{Contrastive Representation Learning}
\label{sec:methodcl}
We apply contrastive learning to refine the semantic space learned by our model so that the representation of the same semantic class entities are pulled closer while the representation of different semantic class entities are pushed outward.

To do this, we firstly generate positive/negative entities for each semantic class from seed sets and previous expansion results.
Note that these previous expansion results are from the last iteration, since our expansion framework is an iterative process.
Positive entities $\mathbb{E}_{pos}$ are defined as seed entities or entities that rank higher than a threshold $\text{thr}_{pos}$ in the expanded ranked lists. The entities that lie in a pre-defined interval $(\text{L}_{neg}, \text{U}_{neg})$ of the expanded ranked lists are automatically selected as negative entities $\mathbb{E}_{neg}$. 
\begin{equation}
    \mathbb{E}_{pos} = \left\{e|e \in \mathbb{E}_{seed}\ \ \text{or rank}(e) < \text{thr}_{pos} \right\},
\end{equation}
\begin{equation}
    \label{Equ_Eneg}
    \mathbb{E}_{neg} = \left\{e| \text{L}_{neg} < \text{rank}(e) < \text{U}_{neg} \right\},
\end{equation}
where these thresholds are the hyper-parameters for positive/negative entities selection.
Additionally, it is worth noting that we determine these thresholds based on a reasonable assuming, i.e., hard negative entities would be ranked close to positive entities during the expansion process. 
Therefore, we set the $\mathbb{E}_{neg}$'s lower bound $\text{L}_{neg}$ a little larger than the size of all positive entities to select negative entities in practice. 

Inspired by~\citep{robinson2020hard}, we design the contrastive learning method which can concentrate on hard negative entities for ESE.
Specifically, we initialize our models in the same way as we discuss above, while attaching an auxiliary projection head $\boldsymbol{p}$ on top of the encoder of our model. The projection head $\boldsymbol{p}$ maps the final hidden embedding of the masked entity into a normalized feature vector $\mathbf{z} \in \mathbb{R}^D$, where $D$ is the dimension of output. 
To calculate the contrastive loss, samples from positive/negative entities are paired up to form positive/negative sample pairs: 
\begin{equation}
    \mathbb{P}_{pos} = \left\{ (\mathbf{x}, \mathbf{x}')|ent(\mathbf{x}) \in \mathbb{E}_{pos}, ent(\mathbf{x}') \in \mathbb{E}_{pos} \right\}, 
\end{equation}
\begin{equation}
    \mathbb{P}_{neg} = \left\{ (\mathbf{x}, \mathbf{x}')|ent(\mathbf{x})=ent(\mathbf{x}') \in \mathbb{E}_{neg}\right\}, 
\end{equation}
where $ent(\mathbf{x})$ indicates the entity corresponding to the training sample $\mathbf{x}$.
The contrastive loss is then defined as follow:
\begin{equation}
    loss_{cl} =- \sum_{i=1}^{2N} \log \frac{S_{i}^{+}}{S_{i}^{+} + S_{i}^{-}}, 
\end{equation}
\begin{equation}
    S_{i}^{+} = e^{\mathbf{z}_i^{\top} \cdot \mathbf{z}_{j(i)} / t},
\end{equation}
\begin{equation}
    S_{i}^{-} = \max(\frac{-(2N-2)\cdot \tau^{+}\cdot S_{i}^{+} +  \widetilde{S_{i}^{-}}}{1-\tau^{+}}, e^\frac{-1}{t}),
\end{equation}
\begin{equation}
    \label{equ_theoritically}
    \widetilde{S_{i}^{-}} = \frac{(2N-2)\sum_{k: k \neq i \neq j(i)} e^{(1+\beta)\mathbf{z}_i^{\top} \cdot \mathbf{z}_{k} / t} }{\sum_{k: k \neq i \neq j(i)} e^{\beta \mathbf{z}_i^{\top} \cdot \mathbf{z}_{k} / t}},
\end{equation}
where ${S_{i}^{+}}$/${S_{i}^{-}}$ respectively reflects the similarity between two training samples from the same/different sample pair, and $j(i)$ indicates that the training samples corresponding to indexes $i$ and $j$ can form a positive/negative sample pair, that is, $(\mathbf{x}_i, \mathbf{x}_j) \in  \mathbb{P}_{pos} \cup \mathbb{P}_{neg}$, $N$ is the size of mini-batch,  $\tau^{+}$ is the class-prior probability which can be estimated from data or treated as a
hyper-parameter, $\beta$ is the hyper-parameter controlling the level of concentration on hard negative samples, $t$ is the temperature scaling factor which we set as 0.5 in all our experiments. It is noted that the training process alternates between the prediction loss and contrastive loss. 

\subsection{Model Selection and Ensemble}
\label{sec:methodev}
It is reasonable to hypothesize that a model which has learned more common semantic meaning of a class will output more consistent representation of seed entities from that class. Under this hypothesis, we design a scoring function to estimate a model's expansion performance on a semantic class:
\begin{equation}
    \text{sco}(\theta, cls) = - \frac{\sum_{i}^{M} \sum_{j:i \neq j }^{M} \text{KL}\_\text{Div}(r(e_i), r(e_j))}{M * (M-1)} ,
\end{equation}
\begin{equation}
    r(e) = \frac{1}{|\mathbb{S}_{e}|} \sum_{\mathbf{x} \in \mathbb{S}_{e}} \boldsymbol{f}(\boldsymbol{g}(\mathbf{x} | \theta) | \theta) ,
\end{equation}
\begin{equation}
    M = |\mathbb{E}_{seed}^{cls}|,
\end{equation}
where $\theta$ is parameters of the model, $\mathbb{E}_{seed}^{cls}$ is the set of seed entities of the class, $\mathbb{S}_{e}$ is the set of all samples of entity $e$, $r(e)$ is the probabilistic representation of entity $e$, $\text{KL}\_\text{Div}$ is KL Divergence.

The overall score of a model on a dataset is then defined as the geometric mean of the model's scores on all the classes:
\begin{equation}
    \widetilde{\text{sco}}(\theta) = - \left\lvert \sqrt[N_{cls}]{\prod_{i}^{N_{cls}} \text{sco}(\theta, cls_i)} \right\rvert.
    \label{score_function}
\end{equation}

With this scoring function, we are able to select top-k models $\Theta_{top}$ from multiple models with only information of seed sets of each class. We ensemble these models as follow:
\begin{equation}
    \widetilde{\boldsymbol{f}(\boldsymbol{g}(\mathbf{x}))} = \frac{1}{|\Theta_{top}|} \sum_{\theta \in \Theta_{top}} \boldsymbol{f}(\boldsymbol{g}(\mathbf{x} | \theta)).
\end{equation}

The practical model training process and analysis on model efficiency are described in Appendix~\ref{Appendix_A}.

\begin{algorithm}[t]
    \caption{Window Search}
    \label{alg:Window Search}
    \begin{flushleft}
    \hspace*{0.05in} {\bf Input:}
    candidate entity list $L$; current set $L_{cur}$; window size $w$; anchor distribution $\mathbf{d} \in \mathbb{R}^{V_e}$; entity representation $\mathbf{r} \in \mathbb{R}^{V_e}$; scaling factor $\alpha$; stage step $\tau$; counter $c$. \\
    \hspace*{0.05in} {\bf Output:} 
    target entity $e_{t}$.
    \end{flushleft}
    \begin{algorithmic}[1]
        \State $c \leftarrow 0$;
        \State $\text{s}_t \leftarrow -\infty$;
        \State $p \leftarrow \frac{1}{V_e}$;
        \For{$e$ in $L$}
        \If{$c \geq w$}
        \State \textbf{break}
        \EndIf
        \State $\mathbf{r} \leftarrow \frac{1}{|\mathbb{S}_{e}|} \sum_{x \in \mathbb{S}_{e}} \widetilde{\boldsymbol{f}(\boldsymbol{g}(x))}$;
        \State $\mathbf{d} \leftarrow [p]^{V_e}$;
        \State $\mathbf{d}[index(e)] \leftarrow \mathbf{r}[index(e)]$;
        \For{$i$ in $|L_{cur}|$}
        \State $\mathbf{d}[index(e_i)] \leftarrow p * \alpha * 2^{-\lfloor\frac{i}{\tau}\rfloor}$;
        \EndFor
        \State $\mathbf{d} \leftarrow \text{Softmax}(\mathbf{d})$;
        \State $\text{s}(e) \leftarrow -\text{KL}\_\text{Div} (\mathbf{r},\mathbf{d})$;
        \If{$\text{s}(e) > \text{s}_t$}
        \State $e_{t} \leftarrow e$;
        \State $\text{s}_t \leftarrow \text{s}(e)$;
        \EndIf
        \State $c \leftarrow c + 1$;
        \EndFor
    \State \Return $e_{t}$.
    \end{algorithmic}
\end{algorithm}

\begin{table*}[h]
    \centering
    \scalebox{1.00}{
    \begin{tabular}{lccccccccc}
    \toprule \multirow{2}{*} { \textbf{Methods} } & \multicolumn{3}{c} { \textbf{Wiki} } & \multicolumn{3}{c} { \textbf{APR} } & \multicolumn{3}{c} { \textbf{SE2} } \\ \cmidrule(r){2-4} \cmidrule(r){5-7}\cmidrule(r){8-10}
    & MAP@10 & MAP@20 & MAP@50 & MAP@ 10 & MAP@20 & MAP@50 & MAP@10 & MAP@20 & MAP@50 \\
    \midrule Egoset & 0.904 & 0.877 & 0.745 & 0.758 & 0.710 & 0.570 & 0.583 & 0.533 & 0.433 \\
    SetExpan & 0.944 & 0.921 & 0.720 & 0.789 & 0.763 & 0.639 & 0.473 & 0.418 & 0.341 \\
    SetExpander & 0.499 & 0.439 & 0.321 & 0.287 & 0.208 & 0.120 & 0.520 & 0.475 & 0.397 \\
    CaSE & 0.897 & 0.806 & 0.588 & 0.619 & 0.494 & 0.330 & 0.534 & 0.497 & 0.420 \\
    CGExpan  & \textbf{\underline{0.995}} & \underline{0.978} & 0.902 & \underline{0.992} & \underline{0.990} & 0.955 & 0.601 & 0.543 & 0.438 \\
    SynSetExpan & 0.991 & \underline{0.978} & \underline{0.904} & 0.985 & \underline{0.990} & \textbf{\underline{0.960}} & \underline{0.628} & \underline{0.584} & \underline{0.502}\\
    \midrule \modelName{} & \textbf{0.995} & 0.982 & 0.926 & 0.993 & 0.990 & 0.934 & \textbf{0.683} & \textbf{0.633} & \textbf{0.541} \\
    \modelName-CN & \textbf{0.995} & \textbf{0.983} & \textbf{0.929} & \textbf{1.000} & \textbf{0.996} & 0.955 & - & - & - \\
    \midrule
    \end{tabular}
    }
    \caption{MAP@K(10/20/50) of different methods. The choices of $K$ value are exactly following the previous works~\citep{zhang-etal-2020-empower,shen-etal-2020-synsetexpan}. All baseline results are directly from other published paper. Note that the class name guidance step in CGExpan is proposed for relatively coarse-grained semantic classes, while the semantic classes of SE2 dataset are more fine-grained, so this method is not very operable on SE2 dataset. We underline the previous state-of-the-art performance on three datasets for convenient comparison.}
    \label{tab:allresult}
\end{table*}

\subsection{Probabilistic Entity Set Expansion}
\label{sec:methodexpan}
Our proposed \modelName{} is an iterative framework based on the probabilistic representation of entities and entity sets.
At the beginning of expansion, we initialize the current set $L_{cur}$ as the given seed set.
In every expansion step, we first calculate the probabilistic representation of current set $r(L_{cur})$ with our pre-trained ensemble model:
\begin{equation}
    r(L_{cur}) = \frac{1}{|L_{cur}|} \sum_{e\in L_{cur}} \frac{1}{|\mathbb{S}_{e}|} \sum_{\mathbf{x} \in \mathbb{S}_{e}} \widetilde{\boldsymbol{f}(\boldsymbol{g}(\mathbf{x}))}.
\end{equation}
$r(L_{cur})$ is essentially the average of predicted entity distributions of all entities in current set, whose dimension is the size of the entity vocabulary. Sorting it and filtering out entities in current set give us a ranked candidate entity list $L$.

The window search Algorithm~\ref{alg:Window Search} on $L$ is to expand the target entities of current set. The algorithm judges the quality of a candidate entity by the similarity between its representation $\mathbf{r} \in \mathbb{R}^{V_e}$ and the anchor distribution $\mathbf{d} \in \mathbb{R}^{V_e}$ of current set. 
Therefore, an entity that is not so prominent (i.e., long-tail entity) but shares more similar representation with current set will be expanded in current set. 
The anchor distribution $\mathbf{d}$ reflects entity distribution of current set, where seed entities and entities expanded earlier weigh heavier. The base of it is set as $\frac{1}{V_e}$, the average entity prediction probability. To make the anchor distribution robust to candidate entities, the anchor probability of candidate entity is set to be the same as the predicted probability of candidate entity. And the anchor probability of each entity in current set scales over $p$, where entities with higher ranks will get larger scale. Note that the anchor distribution $\mathbf{d}$ is transformed into a probability distribution by Softmax before calculating the $\text{KL}\_\text{Div}$.

We increase window size $w$ according to the current set size, since the anchor distribution will be more concrete as the current set size grows larger:
\begin{equation}
    w = w_0 + g * \lfloor \frac{|L_{cur}|}{s} \rfloor,
\end{equation}
where $w_0$ is the initial window size, $g$ is window growing rate, $s$ is window growing step.

Once expanded set reaches target size $\text{S}_{tgt}$, we stop the expansion and run the entity re-ranking algorithm. In particular, for every entity $e$ in the expanded set, we first calculate its score $\text{s}(e)$ in the same way as we do in the window search algorithm. A ranked list $L_{rank}$ can be constructed according to these scores. The aggregation score of every expanded entity is then calculated as follow:
\begin{equation}
    score(e_i) = \sqrt{\frac{1}{i} * \frac{1}{rank(e_i)}}, \quad i = 1 ... \text{S}_{tgt},
\end{equation}
where $i$ is the expand order of entity $e_i$ in expanded set, $rank(e_i)$ is the rank of entity $e_i$ in $L_{rank}$. 

Sorting the expanded set according to these aggregation scores will get the final expansion results.

\section{Experiments}
\label{sec:exp}

\subsection{Experiment Setup}
\label{sec:ExperimentSetup}

\noindent\textbf{1. Datasets.} To verify the correctness of our intuition and proposed method, we choose two public datasets widely used in previous work and an additional recently released larger and more challenging dataset~\citep{shen-etal-2020-synsetexpan}: 
\begin{enumerate}
    \item \textbf{Wiki} and \textbf{APR}, which contains 8 and 3 semantic classes respectively. Each semantic class has 5 seed sets and each seed set has 3 queries, following the previous work.
    \item \textbf{SE2}, which contains 60 semantic classes and 1200 seed queries. The scale of dataset shows that SE2 is more challenging. The datasets used in the experiment are detailed in Appendix~\ref{Appendix_B}.
\end{enumerate}

\noindent\textbf{2. Compared methods.} 
We will compare the following ESE methods in our experiments, the implementation details and hyper-parameter choices of our experiments are shown in Appendix~\ref{Appendix_C}: 
\begin{enumerate}
    \item \textbf{Egoset}~\citep{Egoset}: A multifaceted set expansion system based on skip-gram features, word2vec embeddings and WikiList.
    \item \textbf{SetExpan}~\citep{shen2017setexpan}: A method iteratively selects context features from the corpus and proposes an ensemble mechanism to rank entities.
    \item \textbf{SetExpander}~\citep{mamou-etal-2018-term}: A corpus-based model for expanding a seed entity set into a more complete entity set that belong to the same semantic class.
    \item \textbf{CaSE}~\citep{CaSE}: A framework that constructs candidate entities with lexical features and ranks candidates using the similarity of distributed representation. 
    \item \textbf{CGExpan}~\citep{zhang-etal-2020-empower}: A method that generates the target semantic class name by querying a pre-trained language model and utilizes generated class names to expand new entities. 
    \item \textbf{SynSetExpan}~\citep{shen-etal-2020-synsetexpan}: Current state-of-the-art method that jointly conducts two related tasks and utilizes synonym information to improve performance of ESE.
    \item \textbf{\modelName}: In our proposed framework, we first apply contrastive learning on entity representation model to obtain better entity semantic representation. Then we use model selection and ensemble to avoid the randomness of the pre-training process. Finally we run two novel algorithms to get expansion results.
    \item \textbf{\modelName-CN}: Because our proposed entity representation model is end-to-end trainable, we can combine it with the class name guidance step in CGExpan. 
\end{enumerate}

\noindent\textbf{3. Evaluation Metrics.} The task objective of ESE is to expand a ranked list of entities belong to the same semantic class. Thus, to evaluate the ranked result, we choose to use the \textbf{Mean Average Precision at different top K positions} as: MAP@K $=\frac{1}{|Q|} \sum_{q \in Q} \mathrm{AP}_{K}\left(L_{q}, S_{q}\right)$, where $Q$ is the set of all seed queries and for each query $q$, we use $\mathrm{AP}_{K}\left(L_{q}, S_{q}\right)$to denote the traditional average precision at position $K$ given a ranked list of entities $L_q$ and a ground-truth set $S_q$. To ensure the fairness of experiment, we are completely consistent with the baseline methods' evaluation metric settings. 

\subsection{Experiment Results}
\label{sec:ExperimentResults}
We will first report the overall performance, then analyze and explain the experiment results comprehensively. 

\noindent\textbf{1. Overall Performance.} Table~\ref{tab:allresult} shows the overall performance of different ESE methods. We can see that \modelName{} along with its variant outperform all baselines including current state-of-the-art methods on three datasets, which demonstrates the effectiveness of our proposed method. It is also worth noting that the Wiki and APR are small and relatively easy, the baselines 
don't leave us much space for improvement. But even so, our methods still perform well compared to the baselines.

\begin{table}[h]
\centering
\scalebox{1.00}{
\begin{tabular}{cc}
\hline \textbf{Semantic Class} & \textbf{MAP@100} \\
\hline China Provinces & 0.824 - 0.728 = 0.096 \ \ $\uparrow$\\
Companies & 0.969 - 0.950 = 0.019 \ \ $\uparrow$\\
Countries & 0.930 - 0.941 = -0.011 \ \ $\downarrow$ \\
Disease & 0.959 - 0.948 = 0.011 \ \ $\uparrow$\\
Parties & 0.948 - 0.913 = 0.035 \ \ $\uparrow$\\
Sports Leagues & 1.000 - 0.909 = 0.091 \ \ $\uparrow$ \\
TV Channels & 0.888 - 0.875 = 0.013 \ \ $\uparrow$ \\
US States & 0.763 - 0.750 = 0.013 \ \ $\uparrow$ \\
\hline \textbf{Overall} & \textbf{0.033} \ \ $\uparrow$\\
\hline
\end{tabular}
}
\caption{The improvement (MAP@100) of \modelName{} based on CGExpan under different classes.}
\label{tab:fineresult}
\end{table}

\noindent\textbf{2. Performance Analysis.} (1) For different datasets, our methods stably perform at a competitive level while existing methods fluctuate fiercely. Especially on SE2, which has more entities and semantic classes, our model's advantage is more obvious. (2) For different semantic classes, Table~\ref{tab:fineresult} shows that \modelName{} outperforms previous work under most classes, even though we use more challenging evaluation metric such as MAP@100. (3) For flexibility and expandability, the performance improvement of \modelName-CN compared with \modelName{} suggests that our proposed method can be combined with other methods friendly.  

\begin{figure}[tp]
\centering
\subfigure[Wiki Dataset-$\text{L}_{neg}$] { \label{wikil} 
\includegraphics[height = 0.40 \columnwidth,width=0.46\columnwidth]{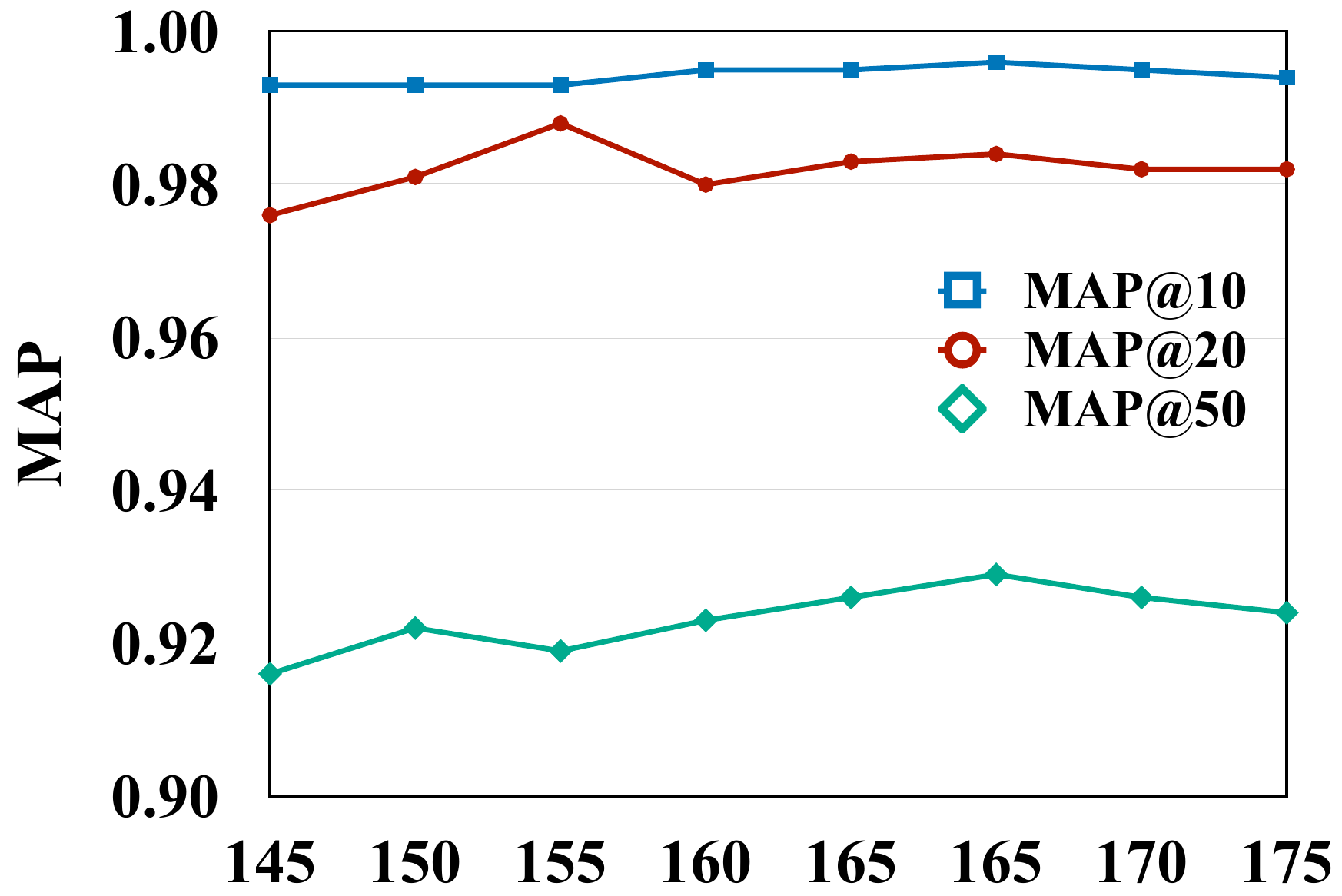} 
} 
\subfigure[Wiki Dataset-$\text{U}_{neg}$] 
{ \label{wikiu} 
\includegraphics[height = 0.40 \columnwidth, width=0.46\columnwidth]{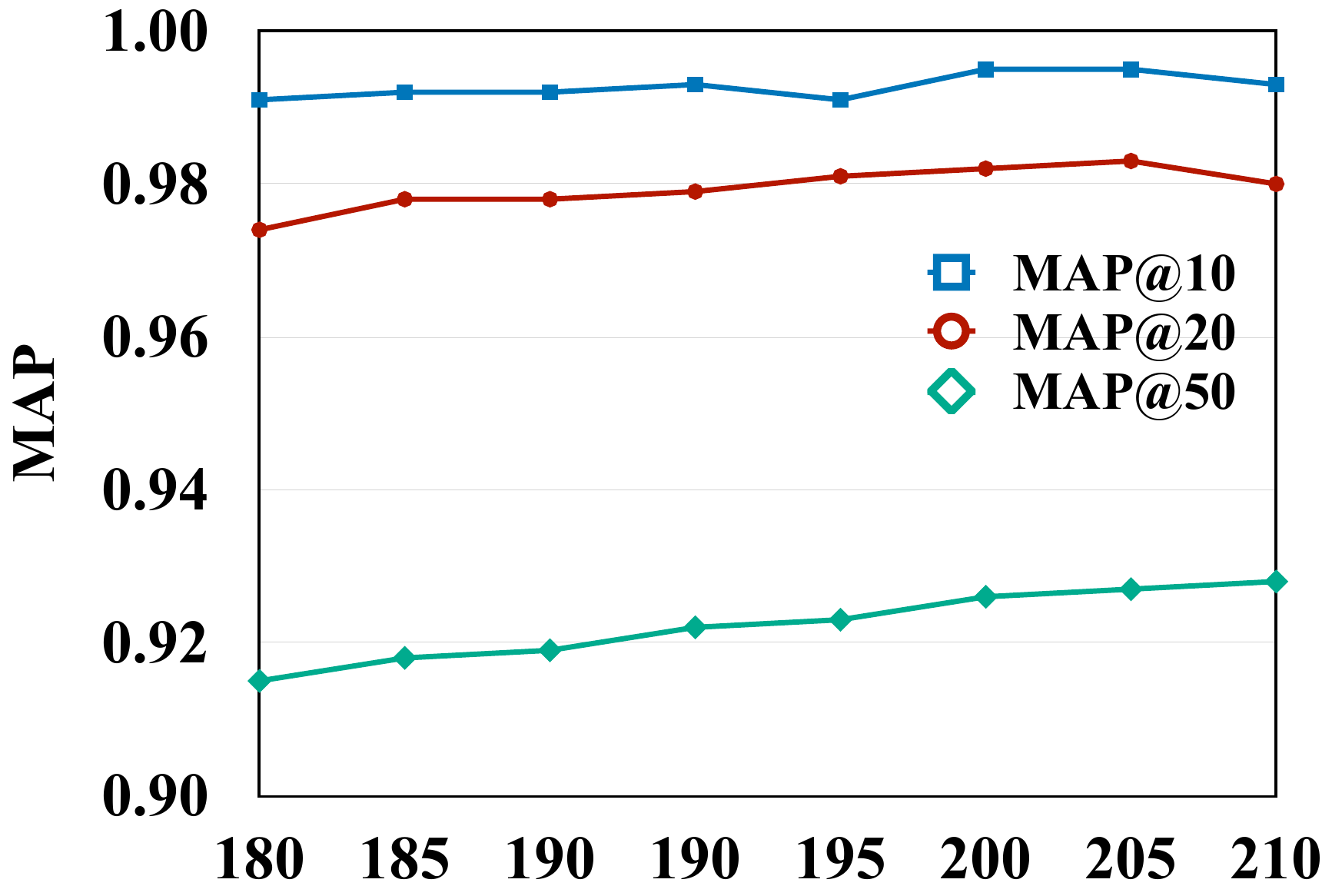} 
} 
\subfigure[APR Dataset-$\text{L}_{neg}$] 
{ \label{aprl} 
\includegraphics[height = 0.40 \columnwidth, width=0.46\columnwidth]{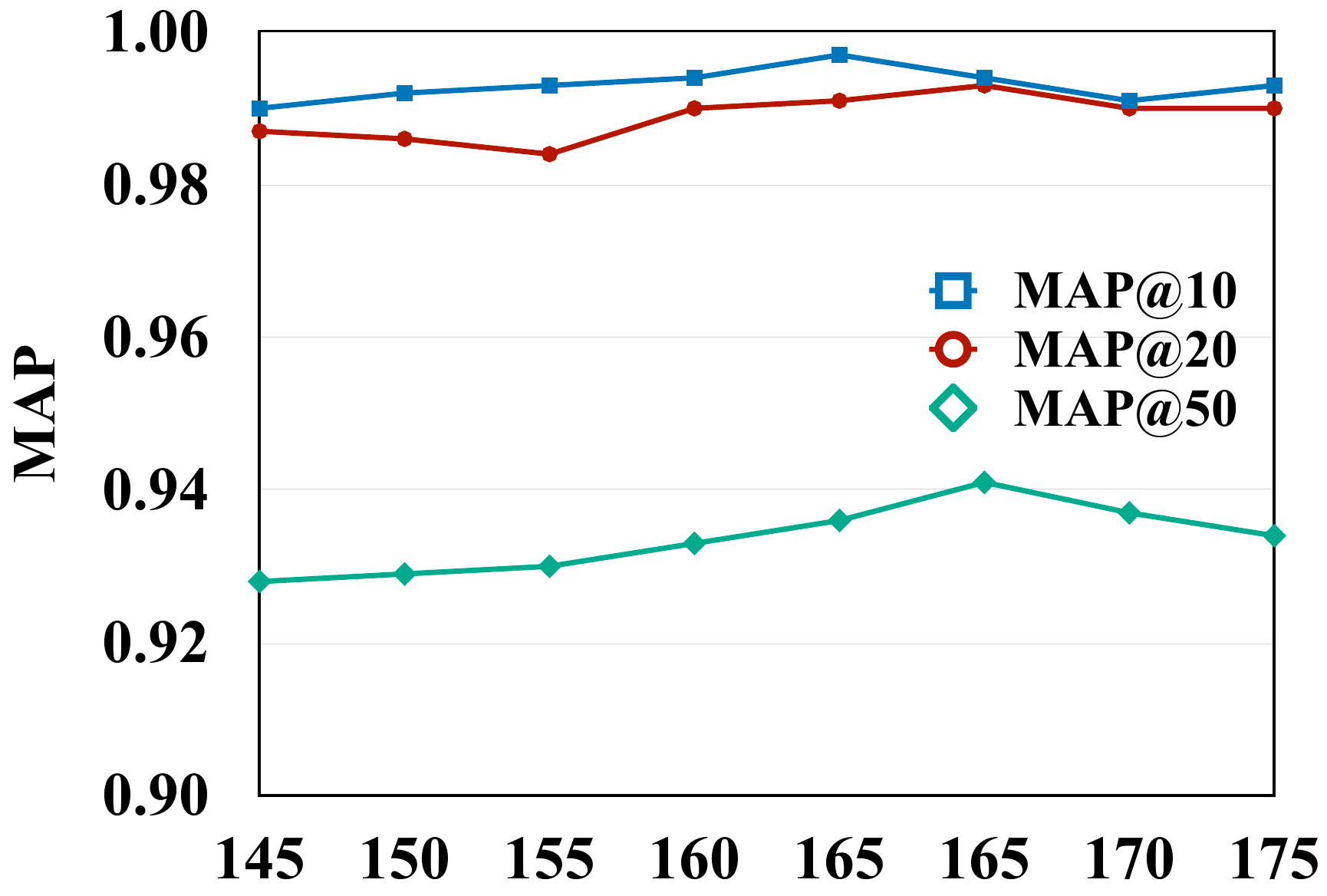} 
}
\subfigure[APR Dataset-$\text{U}_{neg}$] 
{ \label{apru} 
\includegraphics[height = 0.40 \columnwidth, width=0.46\columnwidth]{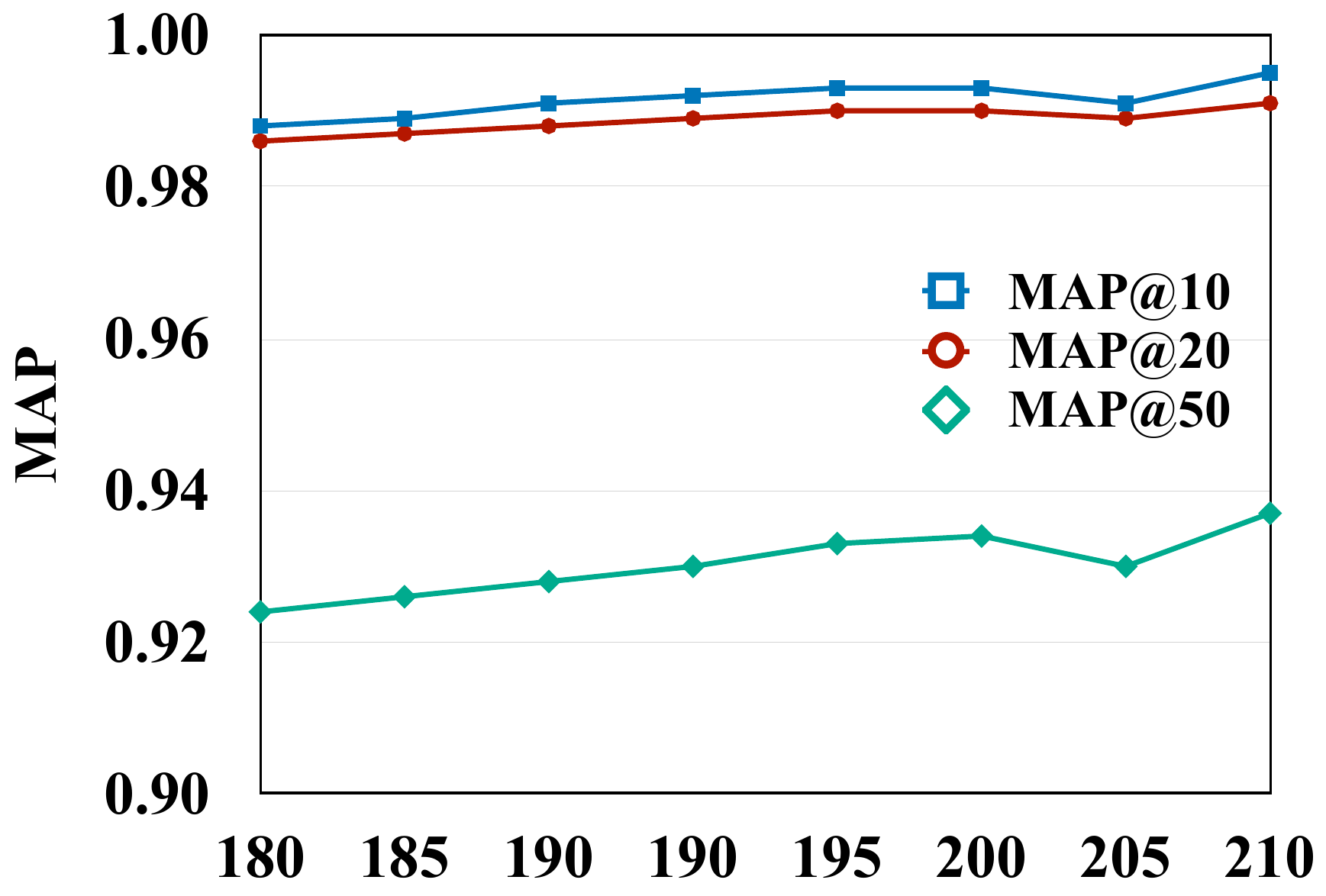} 
} 
\subfigure[SE2 Dataset-$\text{L}_{neg}$] 
{ \label{se2l} 
\includegraphics[height = 0.40 \columnwidth, width=0.46\columnwidth]{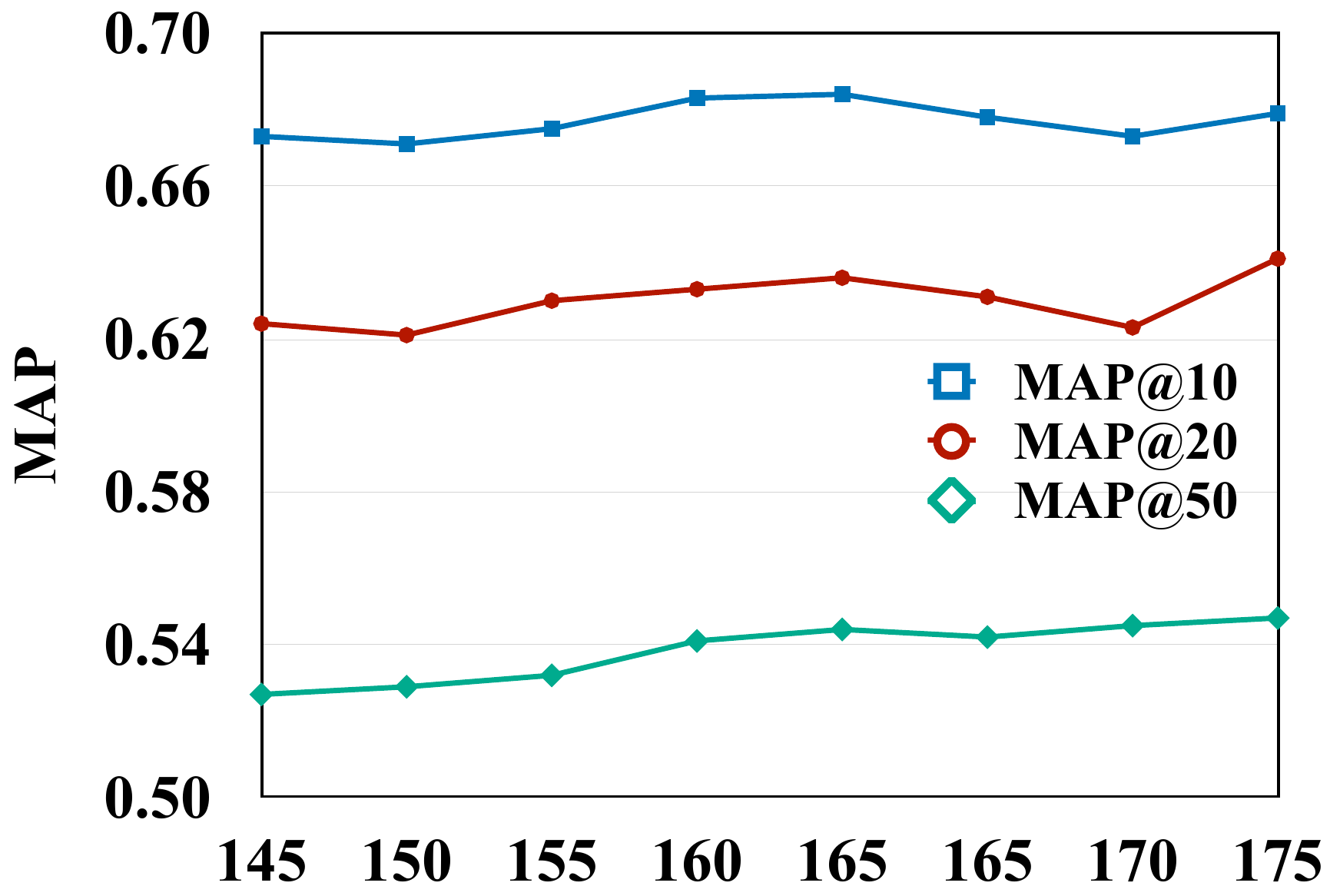} 
} 
\subfigure[SE2 Dataset-$\text{U}_{neg}$] 
{ \label{se2u} 
\includegraphics[height = 0.40 \columnwidth, width=0.46\columnwidth]{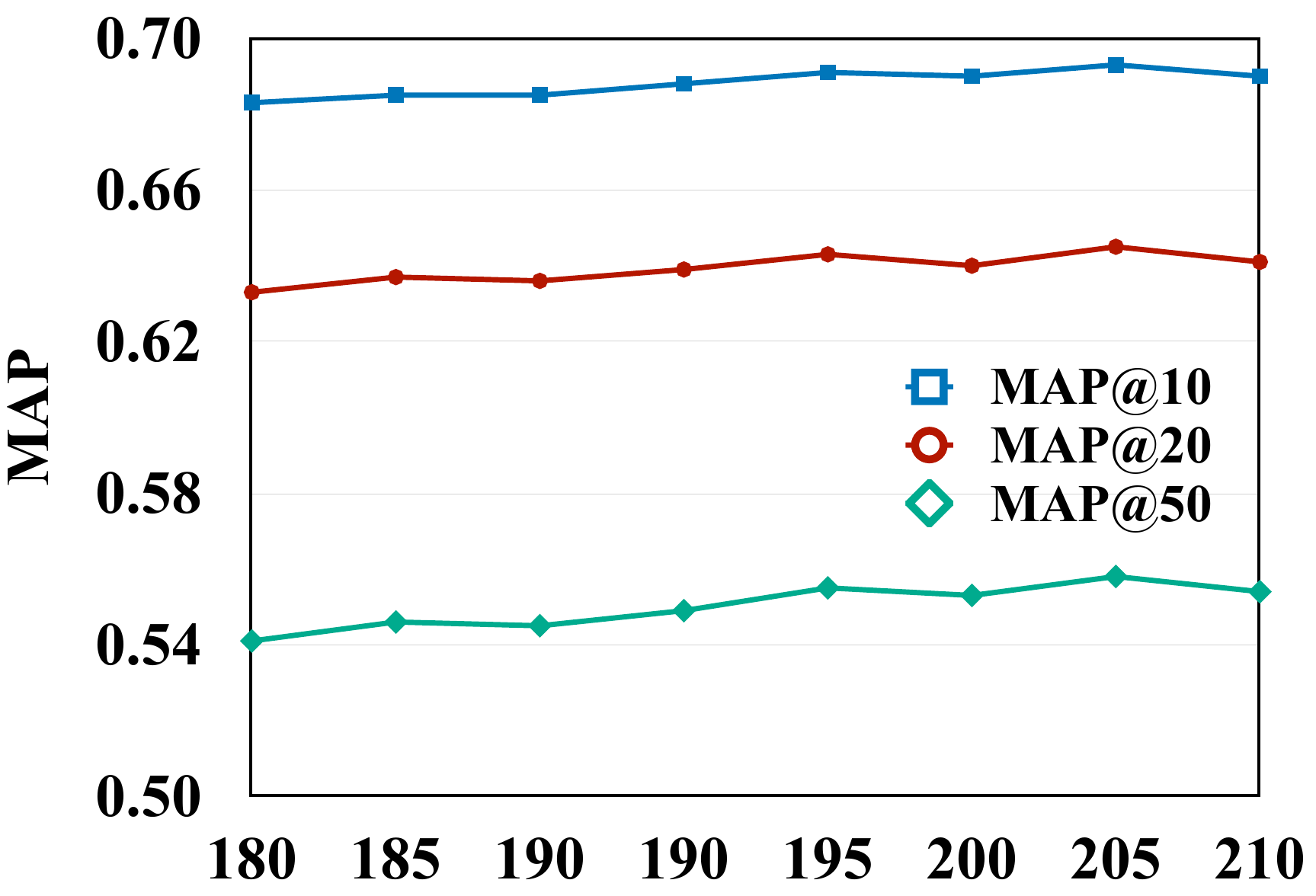} 
} 
\caption{Sensitivity analysis of $\text{L}_{neg}$ / $\text{U}_{neg}$ in \modelName{}.} 
\label{sensitive} 
\end{figure} 

\begin{table*}[h]
    \centering
    \scalebox{1.10}{
    \begin{tabular}{lcccccc}
    \toprule \multirow{2}{*} { \textbf{Methods} } & \multicolumn{3}{c} { \textbf{Wiki} } & \multicolumn{3}{c} { \textbf{APR} } \\  \cmidrule(r){2-4} \cmidrule(r){5-7}
    & MAP@10 & MAP@20 & MAP@50 & MAP@ 10 & MAP@20 & MAP@50 \\
    \midrule CGExpan-NoCN & 0.968 & 0.945 & 0.859 & 0.909 & 0.902 & 0.787 \\
    \modelName{}-NoCLEN & 0.983 & 0.974 & 0.910 & 0.990 & 0.977 & 0.898 \\
    \modelName{}-NoEN & 0.989 & 0.980 & 0.899 & 0.992 & 0.981 & 0.912 \\
    \modelName{}-NoCL & 0.991 & 0.980 & 0.917 & \textbf{0.993} & 0.984 & 0.910 \\
    \modelName{} & \textbf{0.995} & \textbf{0.982} & \textbf{0.926} & \textbf{0.993} & \textbf{0.990} & \textbf{0.934} \\
    \bottomrule
    \end{tabular}
    }
    \caption{Ablation studies of \modelName{} and its variants on two datasets. We arrange the results from top to bottom in the order of increasing components of the model.}
    \label{tab:ablationresult}
\end{table*}

\subsection{Parameter Studies}
In Section~\ref{sec:methodcl}, we propose to automatically select negative entities using a pre-defined interval $(\text{L}_{neg}, \text{U}_{neg})$, according to the Equation~\ref{Equ_Eneg}. Furthermore, to select those really hard negative entities as accurately as possible, we will manually ensure that the value of $\text{L}_{neg}$ is slightly larger than the size of positive entities when we determine the values of these two hyper-parameters. Therefore, it is reasonable to suspect that the values of $(\text{L}_{neg}$ and $\text{U}_{neg})$ will affect the hardness of the selected negative entities, thereby affecting the performance of the \modelName{}. But we can prove both theoretically and empirically that such a phenomenon that parameters affect performance does not exist in our proposed framework.

\textbf{Theoretically}, even if we set an inappropriate and large $\text{L}_{neg}$, it will not cause a drop in the overall performance of the \modelName{}, because our proposed contrastive loss can adaptively focus on really hard entities in a training batch. The negative entities that are more similar to the positive entities will receive higher weight when calculating loss through Equation~\ref{equ_theoritically}. \textbf{Empirically},  we carry out the parameter studies as shown in Figure~\ref{sensitive} to verify the insensitivity of \modelName{} to these two hyper-parameters. Specifically, we fix one of $(\text{L}_{neg}$ and $\text{U}_{neg})$ and change the value of the other, and run the \modelName{} on different datasets to test its performance. From Figure~\ref{sensitive}, we can see that the performance of our proposed \modelName{} is not very sensitive to their specific values when these two parameters are within a reasonable range, because as $\text{L}_{neg}$ or $\text{U}_{neg}$ changes, the model performance (MAP@K) does not change very significantly. \textbf{To sum up}, the values of $(\text{L}_{neg}$ and $\text{U}_{neg})$ will indeed determine what entities we select as the hard negative entities, but due to the design of other structures and training strategy of our model, their values will not affect the overall performance of the model significantly.

\subsection{Ablation Studies}
\label{sec:AbaltionStudy}
To provide a detailed analysis of how our proposed method works on ESE, we perform a series of ablation experiments to see how each component affects the model's expansion performance. Besides, the ablation results will also provide empirical proofs for our intuitions. 

Because the full method of CGExpan leverages some fixed patterns well manually designed  by researchers(i.e., Hearst patterns~\citep{hearst-1992-automatic}), to ensure ablation studies' fairness, we will compare \modelName's variants with CGExpan-NoCN~\citep{zhang-etal-2020-empower}, which mainly consists of a traditional pre-trained language model such as BERT. The \modelName's variants include: 
\begin{enumerate}
    \item \modelName{}-NoCLEN: The ablation of \modelName{} without contrastive learning and model selection and ensemble.
    \item \modelName{}-NoEN: The ablation of \modelName{} without model selection and ensemble.
    \item \modelName{}-NoCL: The ablation of \modelName{} without contrastive learning. 
\end{enumerate}
The results of these methods are shown in Tabel~\ref{tab:ablationresult}.

\noindent\textbf{1. Can Entity Representation Model Empower ESE?} From Table~\ref{tab:ablationresult} we can see that \modelName{}-NoCLEN has a great improvement compared to CGExpan-NoCN, especially for the MAP@50. The significant improvement of \modelName{}-NoCLEN indicates the entity-level masked language model can represent entities better. Besides, it is worth noting that the \modelName{}-NoCLEN's results on APR 
are better than results on Wiki, which is exactly the opposite of CGExpan-NoCN. Because CGExpan-NoCN incorporates the average $\text{BERT}$ representation to select entities and the $\text{BERT}$ is pre-trained on Wikipedia corpus which is similar to the corpus of Wiki dataset in ESE. Therefore, CGExpan-NoCN cannot handle other source corpus, which also reflects that the entity representation model we design is not sensitive to the source corpus and has good generalization performance. 

\begin{figure} 
\centering
\subfigure[Wiki Dataset] { \label{wiki} 
\includegraphics[scale=0.25]{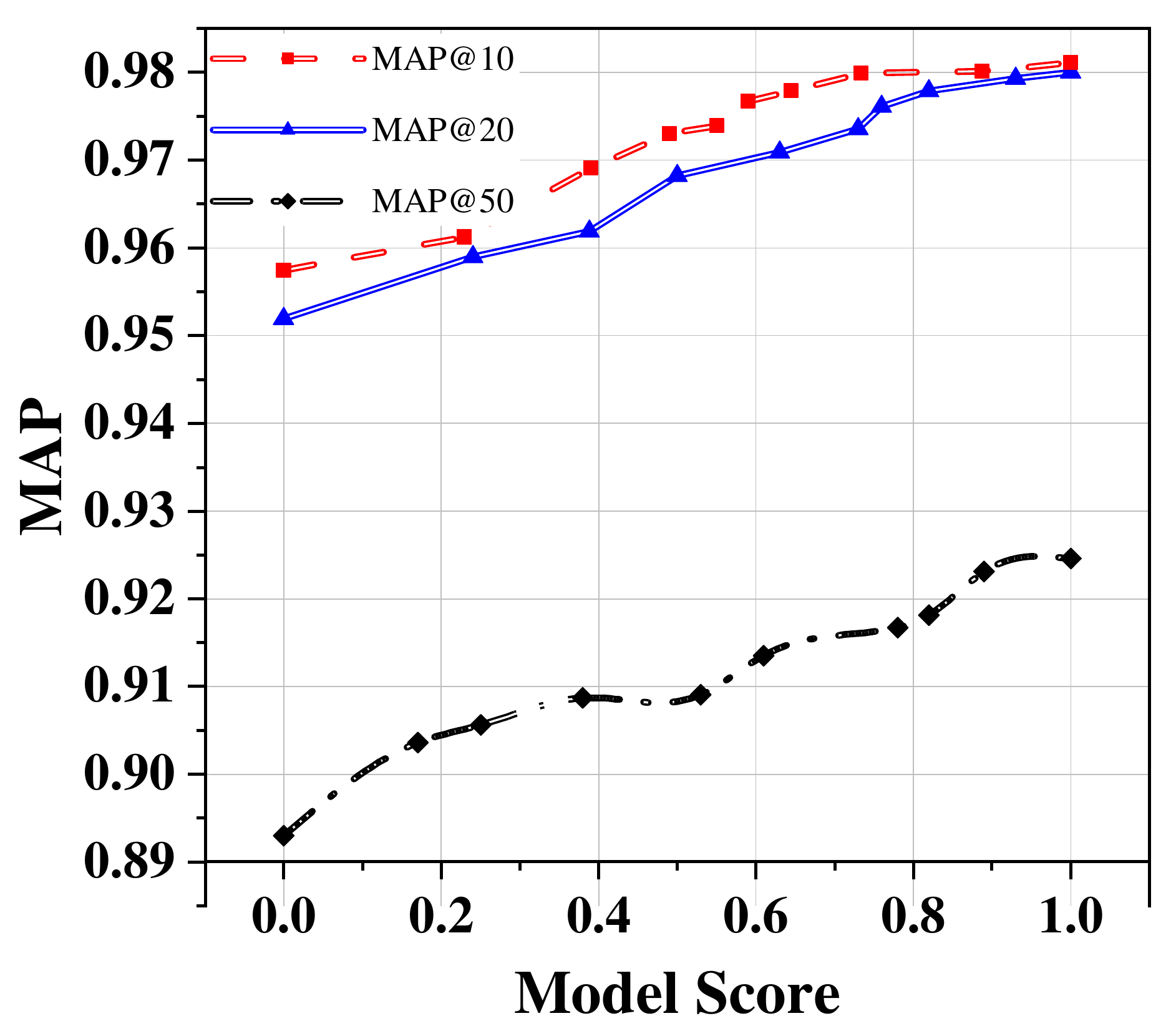} 
} 
\subfigure[APR Dataset] { \label{apr} 
\includegraphics[scale=0.25]{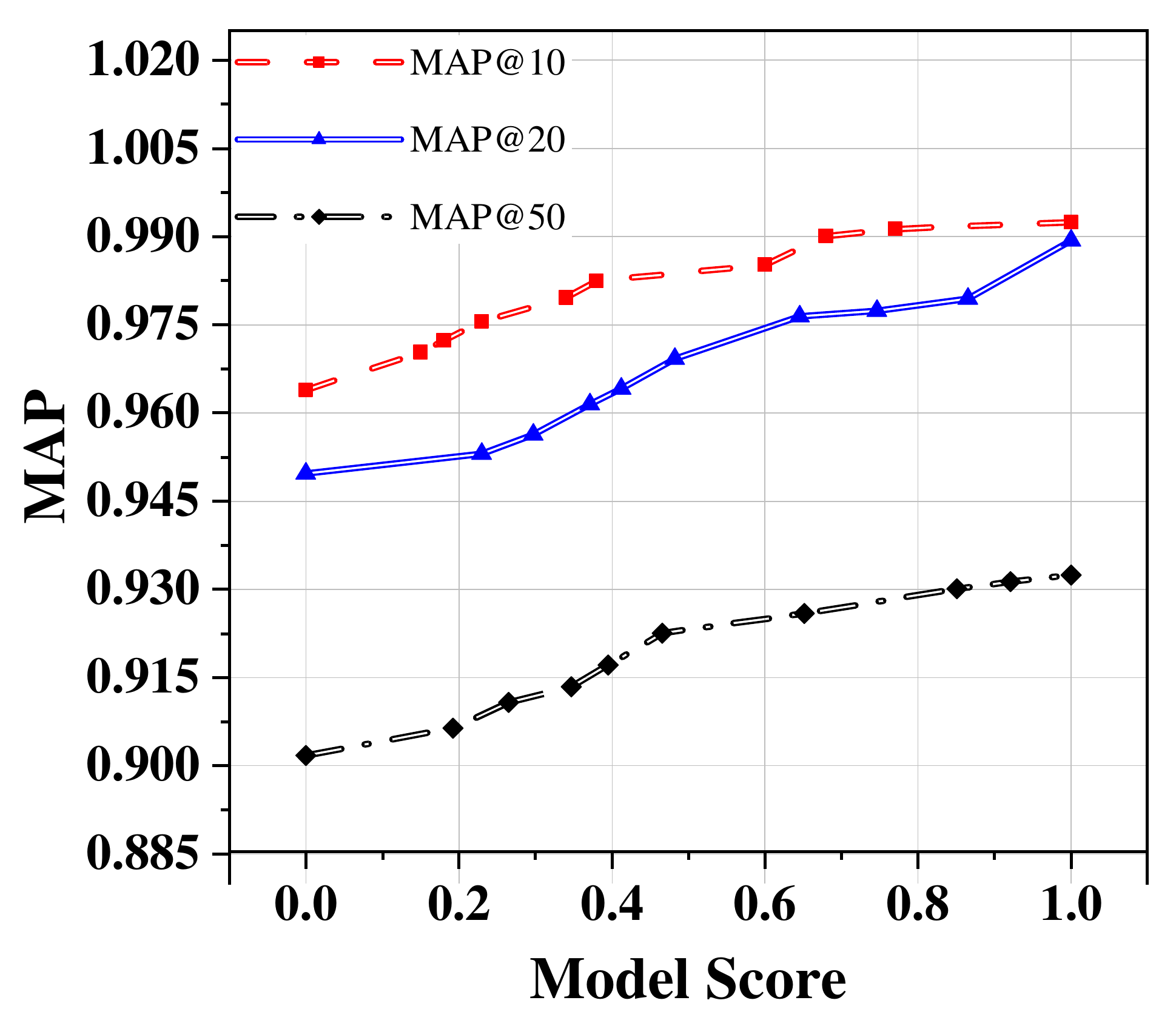} 
} 
\caption{Correlation analysis of model score and performance on Wiki and APR datasets.} 
\label{score_map} 
\end{figure} 

\begin{figure*}[tp]
\centering
\includegraphics[width=1.00\textwidth]{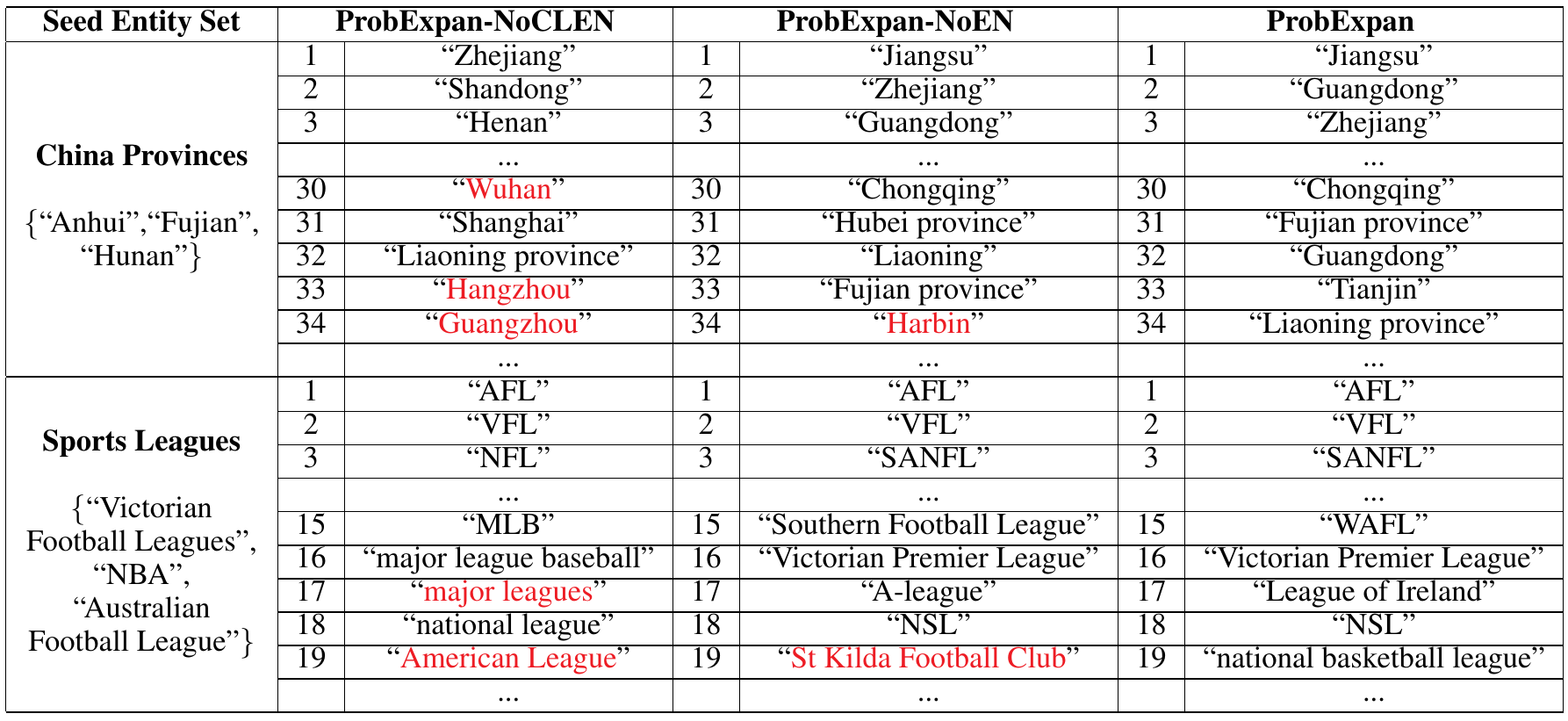}
\caption{Results of two seed entity sets with different semantic classes. We mark the wrong entities in red.}
\label{caseresult}
\end{figure*}

\noindent\textbf{2. Can Contrastive Learning Divide A Clearer Semantic Boundary?} The comparison between \modelName{}-NoEN and \modelName{}-NoCLEN shows that contrastive learning effectively refines the entity representation. According to our observation, previous works such as CGExpan already have competitive performance, the most error-prone case is that they face entities that are semantically ambiguous. This is also the motivation we choose contrastive learning to handle these hard negative entities. The performance results of Table~\ref{tab:ablationresult} and the case study in Figure~\ref{caseresult} together show that contrastive learning can indeed divide a clearer semantic boundary.

\noindent\textbf{3. Can Model Selection And Ensemble Strategy Work?} The results about ensemble method in Table~\ref{tab:ablationresult} show that the model selection and ensemble step we design can bring remarkable improvement. Especially for the \modelName{}'s results, we are pleasantly surprised to find that on the basis of \modelName{}-NoEN, application of model selection and ensemble strategy can still improve further. 
In addition, to verify the validity of the Equation~\ref{score_function}, we analyze the correlation between model score and performance. For the convenience of display, we normalize the model score. The positive correlation results presented in Figure~\ref{score_map} show that the Equation~\ref{score_function} can effectively evaluate the model.

\subsection{Case Studies}
\label{sec:CaseStudy}
We will present different models' representative expansion cases as further verification of our methods' advantages. 

Figure~\ref{caseresult} shows some expansion results of \modelName{}'s  variants for several queries from different semantic classes. We see that even though \modelName{}-NoCLEN has achieved very good overall performance (as can be seen from Table~\ref{tab:ablationresult}), it still occasionally has difficulty distinguishing some hard negative samples. For example, municipal administrative regions such as “\emph{Wuhan}”, “\emph{Hangzhou}”, and “\emph{Guangzhou}” are likely to have great similarities in context with provincial administrative regions such as “\emph{Shanghai}” and “\emph{Zhejiang}” when training a language model, because they all actually belong to \texttt{Location} entities. Therefore, \modelName{}-NoCLEN cannot represent these entities in a more fine-grained manner at the semantic level. As shown in the comparison between \modelName{}-NoCLEN and \modelName{}-NoEN columns of Figure~\ref{caseresult}, \modelName{}-NoEN can recall more entities belonging to the correct target semantic class.
So we can know that contrastive learning can divide a tighter and clearer boundary for the target semantic class through by extending the distance between negative and positive samples and shortening the distance between positive samples in the semantic space. 

From the \modelName{}-NoEN column of Figure~\ref{caseresult}, we can see contrastive learning still can not solve some extreme situations. For example, suppose a person does not have any external background knowledge, then when he/she sees “\emph{St Kilda Football Club}”, he/she must be easy to literally classify it as \texttt{Sports Leagues}. Therefore, we design the model selection and ensemble mechanism to get better expanded entities on the basis of \modelName{}-NoEN and the mechanism's effectiveness can be reflected from the \modelName{} column of Figure~\ref{caseresult}. From the whole Figure~\ref{caseresult} we can know that the effect of \modelName{}-NoEN is better than \modelName{}-NoCLEN, and \modelName{} can be further improved based on \modelName{}-NoEN. Such experimental results are in line with our design expectations.

\section{Conclusions}
\label{sec:cls}
In this paper, we introduce to pre-train an entity-level masked language model with the entity prediction task. Then we firstly empower the ESE model to better handle hard negative entities with contrastive learning task. To utilize our pre-trained entity representation model, we propose the \modelName{}, a novel probabilistic ESE framework that consists of two simple yet effective algorithms, namely window-search and entity re-ranking algorithms. 
In the future, we will further study how to apply our pre-trained ESE model in cross-domain scenarios to better exploit its generalization ability. Combining various domain adaptation methods with our model will be an interesting direction. Moreover, it is also a worthy and promising research direction to study how to automatically measure the hardness of negative entities, so that the really hard negative entities can be better directly selected.

\section{Acknowledgements}
This research is supported by the Shenzhen General Research Project (Grant No. JCYJ20190808182805919) and the 173 program (Grant No. 2021-JCJQ-JJ-0029), National Natural Science Foundation of China (Grant No. 6201101015), Beijing Academy of Artificial Intelligence(BAAI), Natural Science Foundation of Guangdong Province (Grant No. 2021A1515012640), the Basic Research Fund of Shenzhen City (Grant No. JCYJ20210324120012033), National Key R\&D Program of China (No. 2021ZD0112905) and Overseas Cooperation Research Fund of Tsinghua Shenzhen International Graduate School  (Grant No. HW2021008).

\clearpage
\appendix
\label{sec:appendix}

\section{The Training Process of Model in Practice}
\label{Appendix_A}

In practice, the overall training process of entity representation model consists of four phases: 
\begin{enumerate}
    \item In the first phase, we train multiple models in parallel with a masked entity prediction task.
    \item In the second phase, we select top-k models using an algorithm based on the consistency of probabilistic representation of seed set entities, and ensemble them.
    \item In the third phase, we run the expansion procedure and generate positive/negative entities from the expansion results and seed sets, with which we train multiple models in parallel with the combination of contrastive loss and masked entity prediction loss.
    \item In the fourth phase, we select and ensemble models in the same way as we do in the second phase.
\end{enumerate}

For the overall model efficiency, since the bottom layers of encoder $\boldsymbol{g}$ are frozen and all models are trained in parallel, the time cost is close to fine-tuning single BERT. Once training is done, we parallelly use all models to calculate the predicted entity distributions for all samples on the corpus and saved the output into disk, so that models’ output can be easily retrieved during model selection and ensemble. Overall, the framework is efficient enough for downstream applications.

\section{Datasets Used in Experiments}
\label{Appendix_B}
We select Wiki and APR datasets which are used in previous work and an recently proposed more challenging dataset, SE2. We use these datasets following previous works to process the corpus. The download links of these datasets are all available in their original papers~\citep{zhang-etal-2020-empower, shen-etal-2020-synsetexpan}. The details of these datasets are described as follow:
\begin{enumerate}
    \item \textbf{Wiki}, which is from English Wikipedia articles. It mainly contains 8 semantic classes, namely \texttt{China Provinces}, \texttt{Companies}, \texttt{Countries}, \texttt{Diseases}, \texttt{Parties}, \texttt{Sports Leagues}, \texttt{TV Channels} and \texttt{US States}. Each semantic class has 5 queries and each queries has 3 entities as the initial seed entity set.
    \item \textbf{APR}, which is from news articles published by Associated Press and Reuters in 2015. It mainly contains 3 semantic classes, namely \texttt{Countries}, \texttt{Parties} and \texttt{US States}. Each semantic class has 5 queries and each query has 3 entities as the initial seed entities.
    \item \textbf{SE2}, which is from Wikipedia 20171201 dump. It contains 60 major semantic classes and 1200 seed set queries. It is the latest and largest benchmark for ESE.
\end{enumerate}

\section{Implementation Details and Hyper-parameter Choices}
\label{Appendix_C}
\subsection{Implementation Details of Baselines}
If a baseline method has make its code public available, we directly use the open-source code of this method in our experiments. 
But if there is no open-source code for a baseline method, we directly report its performance obtained in other published paper~\citep{shen-etal-2020-synsetexpan}, because the datasets and evaluation metric used in the original paper are completely consistent with these in our experiments.

\subsection{Implementation Details of \modelName{}}
Our proposed entity representation model adopts the Encoder-Decoder architecture, the encoder is a multi-layer bidirectional Transformer~\citep{vaswani2017attention} that follows the $\text{BERT}_\text{BASE}$ setting, and the decoder is a classification head. Specifically, we firstly tokenize the sentences by the WordPiece tokenizer~\citep{wu2016google}.
The input embedding of each token is the sum of its token embedding and position embedding. The sequence of input embedding then passes through 12 stacked bidirectional Transformer blocks with $H=768$ hidden dimensions and 12 self-attention heads. It is noted that we freeze part of the layers in our model when pre-train the entity representation model to utilize the pre-training knowledge of $\text{BERT}$. 

\subsection{Implementation Details of \modelName{}-CN}

The entity probabilistic representation of our model can be easily used in other ESE frameworks. Here we discuss how to combine our entity representation model with the class name guidance step of CGExpan. In the Class-Guided Entity Selection module of CGExpan, score of a candidate entity $e_i$ is formulated as:
\begin{equation}
    score_i = \sqrt{score^{loc}_i * score^{glb}_i},
\end{equation}
where $i$ is the index of the candidate entity in the vocabulary, $score^{loc}_i$ is related to the guidance class name and we leave it alone, $score^{glb}_i$ measures the similarity between the candidate entity and entities of current set in the embedding space and is calculated as follow:
\begin{equation}
    score^{glb}_i = \frac{1}{|\mathbb{E}_s|} \sum_{e \in \mathbb{E}_s} \textbf{cos}(\textbf{V}_{e_i},\textbf{V}_{e}),
\end{equation}
where $\mathbb{E}_s$ is a set of entities sampled from current set, $\textbf{V}_e$ is the averaged word embedding of entity $e$ by BERT, $\textbf{cos}(\cdot)$ is cosine similarity function. Finally, we replace $score^{glb}_i$ with $\widetilde{score}^{glb}_i$:
\begin{equation}
    \widetilde{score}^{glb}_i = \frac{1}{|\mathbb{E}_s|} \sum_{e \in \mathbb{E}_s} \frac{1}{|\mathbb{S}_e|} \sum_{x \in \mathbb{S}_e} \widetilde{\boldsymbol{f}(\boldsymbol{g}(x)})[i].
\end{equation}

\subsection{Hyper-parameter Choices}
In our experiments, our model is trained by an AdamW optimizer with the betas of (0.9, 0.999), epsilon of $1\mathrm{e}-6$, and weight decay of $1\mathrm{e}-2$. For different datasets, 
the choices of other hyper-parameters are shown in Table~\ref{tab:hyperparameter}.
It is worth noting that we use the standard and general hyper-parameter selection strategy, i.e., grid search the hyper-parameters by model performance on a validation set that does not overlap with the test set.

\begin{table}[h]
    \centering
    \scalebox{0.90}{
    \begin{tabular}{cccc}
    \hline \textbf{Hyper-parameter} & \textbf{Wiki} & \textbf{APR} & \textbf{SE2} \\
    \hline Frozen layers & 11  & 11 & 10 \\
    Initial $lr_{pred}$ & $1 \mathrm{e}-5$ & $1 \mathrm{e}-5$ & $2.5 \mathrm{e}-6$ \\
    Smoothing factor $\eta$ & 0.075 & 0.1 & 0.15 \\
    Initial $lr_{cl}$ & $1.5 \mathrm{e}-5$ & $1.5 \mathrm{e}-5$ & $3.5 \mathrm{e}-6$ \\
    $\text{thr}_{pos}$ & 12 & 10& 5\\
    $\text{L}_{neg}$ & 170 & 175& 160\\
    $\text{U}_{neg}$ & 200 & 200& 180\\
    $\tau^+$ & 0.05 & 0.1 & 0.01 \\
    $\beta$ & 1 & 1 & 2 \\
    \hline
    \end{tabular}
    }
    \caption{The hyper-parameter settings in experiments.}
    \label{tab:hyperparameter}
\end{table}

\bibliographystyle{ACM-Reference-Format}
\bibliography{sample-base}

\end{document}